\documentclass[10pt,twocolumn,letterpaper]{article}
\pdfoutput=1
\usepackage{iccv}
\usepackage{times}
\usepackage{epsfig}
\usepackage{graphicx}
\usepackage{amsmath}
\usepackage{amssymb}
\usepackage{caption}
\usepackage{dblfloatfix}
\usepackage{array}
\usepackage{caption}
\usepackage{graphicx}
\usepackage{graphics}
\usepackage{tabto}

\usepackage[breaklinks=true,bookmarks=false]{hyperref}

\iccvfinalcopy 

\def\iccvPaperID{9204} 

\ificcvfinal\fi

\begin{document}

\title{Real-time Action Recognition for Fine-Grained Actions and The Hand Wash Dataset}

\author{
Akash Nagaraj, Mukund Sood, Chetna Sureka and Gowri Srinivasa\\
\\
Department of Computer Science,\\
PES University, Bangalore, India\\
{\tt\small [akashn1897, mukundsood2013, chetnasureka1501]@gmail.com, gsrinivasa@pes.edu}}

\maketitle
\ificcvfinal\thispagestyle{empty}\fi

\begin{abstract}In this paper we present a three-stream algorithm for real-time action recognition and a new dataset of handwash videos, with the intent of aligning action recognition with real-world constraints to yield effective conclusions. A three-stream fusion algorithm is proposed, which runs both accurately and efficiently, in real-time even on low-powered systems such as a Raspberry Pi. The cornerstone of the proposed algorithm is the incorporation of both spatial and temporal information, as well as the information of the objects in a video while using an efficient architecture, and Optical Flow computation to achieve commendable results in real-time. The results achieved by this algorithm are benchmarked on the UCF-101 as well as the HMDB-51 datasets, achieving an accuracy of 92.7\% and 64.9\% respectively. An important point to note is that the algorithm is novel in the aspect that it is also able to learn the intricate differences between extremely similar actions, which would be difficult even for the human eye. Additionally, noticing a dearth in the number of datasets for the recognition of very similar or fine-grained actions, this paper also introduces a new dataset that is made publicly available, the \textit{Hand Wash Dataset} with the intent of introducing a new benchmark for fine-grained action recognition tasks in the future.
\end{abstract}

\section{Introduction}

In the recent few years, Action Recognition in videos is a topic of research that has received significant attention from the research community. Image Processing algorithms are often extended and adapted to deal with videos. Although progress in the field of video processing is closely linked to advances in the field of image processing, recently we see far fewer advances in the field of video processing when compared to image processing. \par

Video Processing has seen advances using numerous approaches; modeling more extended temporal sequences \cite{donahue2015long, yue2015beyond} incorporating optical flow \cite{simonyan2014two, feichtenhofer2016convolutional}, and learning spatio-temporal filters \cite{tran2015learning, karpathy2014large}, to name a few. However, one of the main factors hindering the progress of action recognition in videos is the dearth of datasets to train on, as most of the publicly available datasets are too noisy, and not representative of real-world application of action recognition in videos. \par

The primary difference between Image Processing and Video Processing is the additional information video data can convey in the time domain, namely, temporal information. The proposed algorithm in this paper is a modification of the two-stream fusion approach described in \cite{simonyan2014two}. The algorithm takes into account the temporal information present in video data using Optical Flow Vectors between particular frames to ascertain both the direction and magnitude of motion in the images, while simultaneously masking the background to an extent.  \par

\begin{figure}[b]
    \centering
    \captionsetup{justification=centering}
    \includegraphics[width=\linewidth]{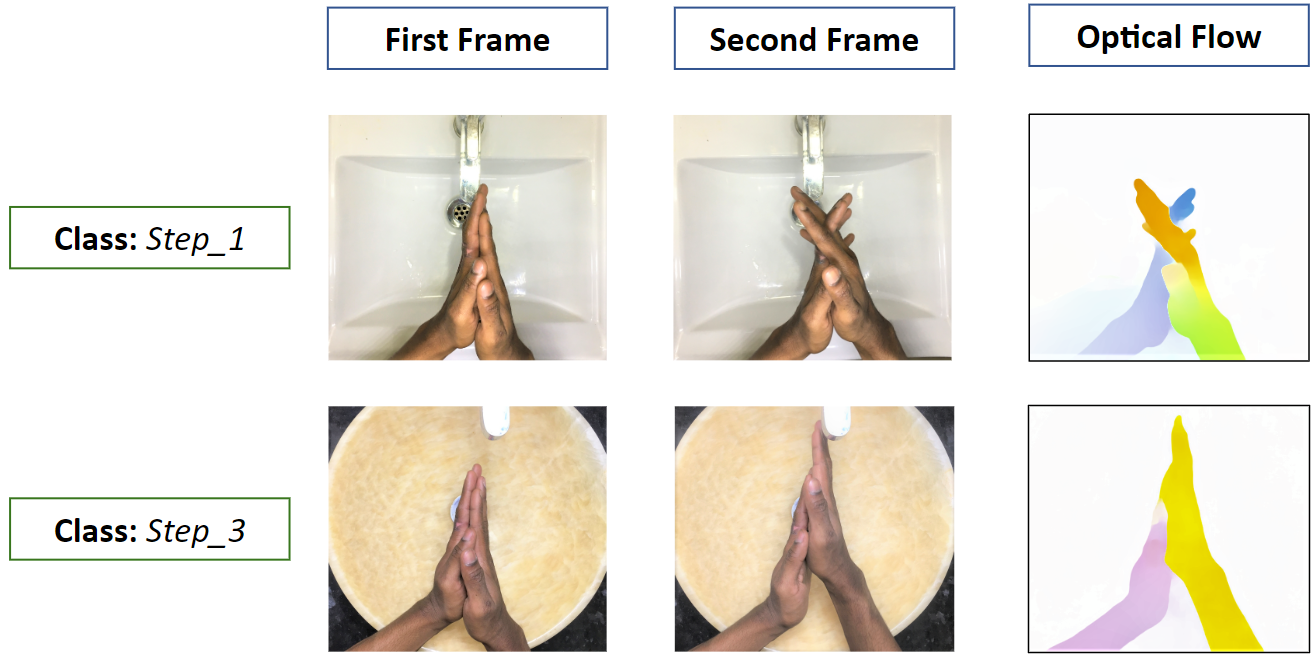}
    \caption{Importance of Temporal Information in Action Recognition with examples from the \textit{Hand Wash Dataset}}
    \label{fig:temporal}
\end{figure}

Most actions could be distinguished using only spatial information, for instance, actions such as swimming, and typing can very easily be distinguished from a single frame of a video of the two actions. However, to understand the importance of the temporal information, consider the actions of closing a door, and opening a door. It is close to impossible to tell if the action being performed in the video is opening a door or closing a door from a single frame of the videos of the two actions. Another good example would be the commonly used example of running and jogging, again, it is nearly impossible to tell the actions apart from single frames. \par

As seen in Fig. 1, consider another example of the need for temporal information from the \textit{Hand Wash Dataset}. Two classes of the dataset, namely \textit{Step\_1} and \textit{Step\_3} are indistinguishable from only Frame 1 of the two actions, but when temporal information, coupled with information of the objects in the videos is taken into account, it is far easier to distinguish the two classes of actions.\par

This is where temporal information provided by data in the video is important. Temporal information tells us about the motion of the subject in a video. The algorithm described in this paper considers the temporal information in two dimensions:
\begin{enumerate}
    \item Recognizing the direction and magnitude of motion between frames.
    \item Identifying how this motion evolves.
\end{enumerate}

In order to capture the temporal information, the algorithm uses Optical Flow Vectors \cite{horn1981determining}. The optical flow vectors between frames are used to recognize the direction and magnitude of motion, while a series of optical flow vectors captures the motion over a predefined time period. \par


The objective of this paper is twofold: (i) to develop a new three-stream fusion algorithm for real-time action recognition in videos that works even on common hardware, and (ii) introduce the \textit{Hand Wash Dataset}. \par


\begin{figure*}[t]
    \centering
    \captionsetup{justification=centering}
    \includegraphics[width=13cm]{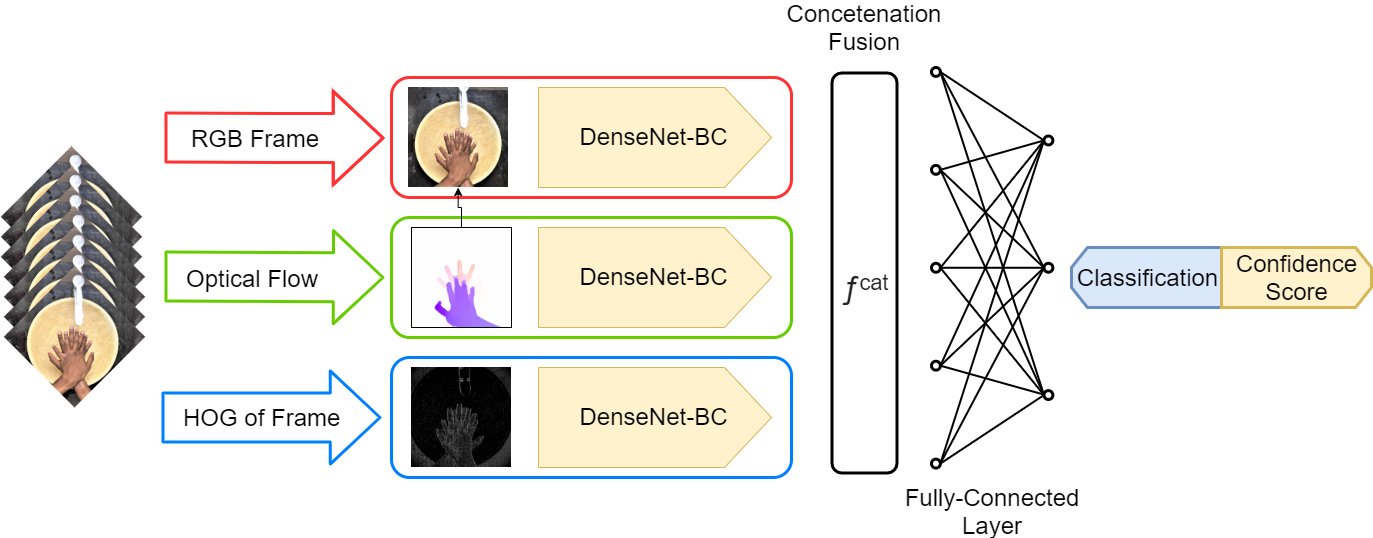}
    \caption{High-level design of the Three-Stream Algorithm}
    \label{fig:hld}
\end{figure*}

\section{Related Work}
Considerable research has been done in the field of action recognition, from the early days of adapting and extending image recognition approaches, to far more complex albeit accurate approaches in the recent years. 

In \cite{sharma2015action}, a multi-layered Recurrent Neural Network (RNN) along with Long Short-Term Memory (LSTM) is used for spatial and temporal deep learning. The model selectively learns parts of the model and can classify after passing a few continuous frames to it by prioritizing between frames in the learning stage to rank them in order of importance for the decision making. For this, the author makes use of the UCF-101 \cite{soomro2012ucf101}, HMDB-51 \cite{kuehne2011hmdb} and Hollywood datasets. A feature cube is obtained after the CNN processes the image which comprises of features differently and spatially located. The model then computes $x_t$ as the dot product of the feature cube $x_t$, $i$ and the SoftMax of the location $l_t$. At a predefined time, $t$, the RNN takes a slice of the feature $x_t$ generated by the dot product and passes it through 3 layers of the LSTM to predict the next class of actions and corresponding possible location probabilities. \\

Exploring the vast amount of research in action recognition also led us to the niche field of anomaly detection in videos, such as \cite{medel2016anomaly}. The authors of \cite{medel2016anomaly} initially define what they believe an anomaly is in a video, and further go on to discuss its implications and possible use cases. Anomalies detected can be used in surveillance situations, intrusion detection situations, health monitoring situations of patients on life-support or who require intensive care or even event detection in a variety of situations. 
They also highlight the difference and difficulty with anomaly detection when it comes to the field of action recognition because actions are very vaguely defined and often cover a wide range of activities.
The authors describe and try to implement a very interesting model, wherein they attempt to train their Convolutional Long-Short Term Memory Network to predict future frames, based on what it has already seen. 
Their model consists of an encoder and two decoders. To the encoder, they feed a sequence of non-overlapping frames in chronological order across several Convolutional LSTM \cite{hochreiter1997long} layers that span over the predetermined number of time-steps. The output of the final Convolutional Layer \cite{krizhevsky2012imagenet} is used as the encoded output. 
There are two decoders, one to decode the past, and one to decode the future. The future decoder is conditioned, since the past decoder can only have one possible outcome. 
Hence, they can build a model that can effectively predict future frames based on past frames. They were able to obtain a mean-squared error of 14.83 between the original future frame, and the predicted future frame. \\

The authors of \cite{sharma2015action} have used an interesting technique that requires weak-labeling of their dataset. The label given to each video is simply the video name, for example, a video need only be labeled as one of the classes mentioned above, or a regular video. \par
They had an equal number of positive and negative samples that were to be fed to the network. From each video, they take 32 temporal segments in two sets. One is a ‘positive bag’ of segments from the normal videos, and the other is a ‘negative bag’ of segments from the anomaly videos. These segments are then passed to Facebook Research's R(2+1)D architecture to generate features for each frame in a particular video. \par
The output of the feature extraction network is then fed to their network frame by frame to generate instance scores for each ‘bag’ of frames. This way, they are not only able to find out whether a video has an anomaly, but also exactly when the anomaly occurred. \\

Reference \cite{diba2017temporal} introduces a new architecture that combines temporal information across variable depth, using a technique to supervise transfer learning between 2 Dimensional Networks to 3 Dimensional Networks. Frames from videos are input into the 2 Dimensional pre-trained Network and Temporal 3-Dimensional Network, but the catch is that the frames could be from a single video or multiple different videos. The architecture learns to predict if the frames are from the same video, or not. Based on this, the error from the prediction is back-propagated through the Temporal 3-Dimensional Network to transfer information effectively. \par
The authors of \cite{carreira2017quo} augment the work done in \cite{diba2017temporal}  but differ in the fact that \cite{diba2017temporal} uses a single stream 3D DenseNet based architecture, using a multi-depth temporal pooling layer, called a Temporal Transition Layer. This is stacked over dense blocks to capture different temporal depths. The concept of pooling with kernels of varying temporal sizes achieves multi-depth pooling. \par

One of the earlier approaches to tackle this problem is seen in \cite{simonyan2014two}. The work done in \cite{simonyan2014two} introduced the two-stream architecture, which laid the foundation for the work presented in this paper in terms of design, but also has its fair share of differences. The authors have achieved two major results; Introducing the two-stream architecture to process videos, which incorporates a spatial stream network, and a temporal stream network, and proving that stacked dense optical flow across multiple frames achieves great results, in spite of a dearth of training data \cite{xiao2006bilateral}.

Reference \cite{simonyan2014two} works on the datasets: UCF-101 \cite{soomro2012ucf101} and HMDB-51 \cite{kuehne2011hmdb}, and as pointed out above, shows that the performance of a multitask learning model performs better on both datasets, with an increase in the training set size.\par
Reference \cite{feichtenhofer2016convolutional} further builds on the work presented in \cite{simonyan2014two}, and focuses mainly on the fusion of the temporal and spatial steams of the two-stream architecture. The authors of \cite{feichtenhofer2016convolutional} experiment with various fusion techniques such as max fusion, bilinear fusion, and concatenation fusion for the spatial stream, and 3D Pooling and 3D Convolution Pooling for the temporal stream. They also focus on where to fuse the two networks and use the VGG-16 architecture for both streams and fusing. The work presented in this paper further builds on the work done in \cite{feichtenhofer2016convolutional}.\par

A common theme seen in almost all the previous methods to solving the problem of action recognition is the approaches are extremely compute-intensive and are far from being used in real-time, or being used in real-world constraints. Another important factor to note is that most of the aforementioned methods have focused on multitask learning of largely different actions (such as Swimming and Typing from UCF-101), but almost none have focused on recognition of fine-grained actions, such as those in the \textit{Hand Wash Dataset}, which would be more representative of a real-world situation where action recognition could potentially be applied.



\section{Approach}
Drawing inspiration from reference \cite{feichtenhofer2016convolutional}, the problem at hand is treated as a video classification problem, as opposed to separating the frames of the video and treating it as an image classification problem like earlier action recognition solutions. The temporal information of the video must also be taken into account, and not only the spatial information. Additionally, from our experiments, we noted that the addition of another stream, with object-level information using the Histogram of Oriented Gradients feature descriptor considerably improved the accuracy of our model while adding an acceptable increase in the number of trainable parameters. This is the reason the algorithm itself consists of three streams, with an efficient deep neural network architecture for each stream; the three streams represent the Spatial Stream of the video, the Temporal Stream, and a stream for only the useful object-level information. A high-level design of this algorithm is seen in Fig. 2.

\subsection{Pre-processing}
Before talking about the classification algorithm, it is important to mention the pre-processing that was performed on the video. The videos used in the training set were all recorded at 30 frames per second.

Firstly, in order to reduce the computational burden, rather than working with all the frames of the video, a fixed number of frames every second are randomly selected to work with. This fixed number is decided by the design parameter; \textit{sample\_frames\_per\_second}. The function randomly selects \textit{sample\_frames\_per\_second} number of frames every second, and pre-processes them, before finally passing them as input to the classification algorithm.
The main pre-processing involved is:
\begin{itemize}
    \item Resizing the frame (112x112 in this case)
    \item Computing Optical Flow \cite{horn1981determining} between the current frame, and the frame next-in-line
    \item Computation of the local object appearance and shape within an image using the HOG \cite{dalal2005histograms} feature descriptor
\end{itemize}

The resizing of the image is performed using bilinear interpolation, with the primary intention being to reduce the computational demand of the classification algorithm as resizing reduces the size of the matrix computations. In our experiments, pre-processing a frame to generate optical flow between two frames took 0.08 seconds per image, while the processing of HOG frames took 0.06 seconds per image on the Intel i5-6500 processor that was used.

\subsection{Temporal Information using Optical Flow}

As mentioned earlier, considering the temporal information of the video, and not only the spatial information is extremely important. To do this, a fast and accurate method of computing optical flow called Coarse2Fine Optical Flow is used. This method of optical flow generation has shown great success in unsupervised learning endeavors. \cite{pathak2017learning} trains Convolution Networks using low-level motion-based grouping, generated by using the Coarse2Fine Optical flow and sees very promising results. A visual representation of Optical Flow is seen in Fig. 3.

The following are a few sets of images for which the optical flow has been generated:

\begin{figure}
    \centering
    \captionsetup{justification=centering}
    \includegraphics[width=8.5cm]{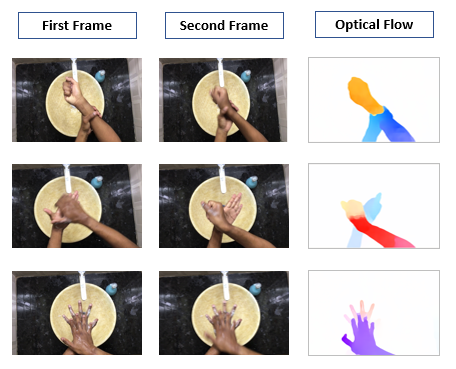}
    \caption{Enhanced Optical Flow generation from frames at time $t-\Delta t$ (first) and at time $t$ (second)}
    \label{fig:opt}
\end{figure}

Reference \cite{feichtenhofer2016convolutional} stacks the optical flows, to form a 10-stack of temporal optical flows to make optimal use of this temporal information of the video, which essentially captures the temporal information in the video in 2 dimensions; one from the difference in motion between a pair of images, and next from stacking these optical flow vectors, to maintain the difference in optical flow between all the frames. \par
Although effective, the three-stream algorithm proposed in this paper does not stack optical flows, but rather uses a single optical flow associated with a frame at time $t - \Delta t$ and time $t$. Since one of the main objectives of the three-stream algorithm is to make action recognition in real-time feasible, the stacking of these frames is very compute-intensive, and would not be possible in real-time. Additionally, the inclusion of the third stream to learn local object data using the HOG feature descriptor, more than makes up for the loss of learning due to the absence of the stacking of optical flows in the temporal stream.

\subsection{Three-Stream Algorithm}
To treat the problem at hand as a video classification problem, as opposed to separating the frames of the video and treating it as an image classification problem, the temporal information of the video must be taken into account, and not only the spatial information. Other approaches that solve the problem of action recognition stack optical flows in the temporal stream, but this is very compute-intensive. \par
We propose a new architecture that includes a stream to learn the local object-level information in the video. This is crucial, as object-level information also gives us information on the source of motion in the video, and how this motion evolves overtime.
DenseNet \cite{huang2017densely} is an architecture that has seen great success in the various domains of Computer Vision. This can be attributed to the fact that it is comprised of \textit{dense blocks}, where the dimensions of the feature maps are constant within a block, while the number of filters changes between blocks. Effectively, using concatenation, every layer receives the collective information from all layers before it. In particular, DenseNet-BC, pre-trained with ImageNet weights is the architecture of choice based on our experiments. DenseNet-BC is a regular DenseNet, but with two modifications:
\begin{itemize}
    \item Bottleneck Layer (B) - DenseNet-BC uses $1x1$ convolutions before $3x3$ convolutions to reduce the size of the feature maps, thereby improving efficiency.
    \item Compression Factor (C) - DenseNet-BC uses a \textit{compression factor} to decrease the number of output feature maps.
\end{itemize}

DenseNet-BC provides accurate results, in near-real time, when used in the three-stream algorithm. Each stream contributes 0.8 million parameters, and a vast majority of the parameters are contributed by the fully-connected layer. The rationale behind using DenseNet-BC over other architectures can be summarized as:
\begin{enumerate}
    \item Smoother decision boundaries - since the classifier uses features of all complexity levels, DenseNet has smoother decision boundaries, and as a consequence, performs better even when there is a dearth of training data.
    \item Strong gradient flow - errors are propagated to the preceding layers directly from later layers.
    \item High feature diversity - as every layer receives the cumulative information of all layers before it, the patterns unearthed are very rich and the features are very diversified.
    \item Compute Efficiency - DenseNet-BC is a deep, and narrow network, which performs accurately, while only having 0.8 million trainable parameters.
\end{enumerate}

In a typical real-world application of the action recognition system, erroneous actions (actions that do not belong to any of the classes in the training data) are very possible. To detect such erroneous actions, we also generate a \textit{confidence score} with each frame prediction. This would be very effective in a real-world application of action recognition, where there are a set of acceptable action classes, and erroneous or anomalous classes can be detected using a combination of the three-stream algorithm, and a frame buffer. This \textit{confidence score} is used with a design parameter, the \textit{threshold confidence} and if the predicted confidence remains below the \textit{threshold confidence} for a stipulated amount of time, the action is classified as an erroneous action.
\\
\subsubsection{Method of Fusion}
The three independent streams are then fused using the concept of Concatenation Fusion.
\begin{equation}
y^{cat} = f^{cat}(x^{a}, x^{b}, x^{c})
\end{equation}

Concatenation is essentially the stacking of the feature maps from the 3 individual steams $i$ and $j$ across the feature channels, denoted by $d$:
\begin{equation}
y^{cat}_{i, j, 3d} = x^{a}_{i, j, d}
\hspace{1.0cm}y^{cat}_{i, j, 3d-2} = x^{b}_{i, j, d}
\end{equation}
\begin{equation}
y^{cat}_{i, j, 3d-1} = x^{c}_{i, j, d}
\end{equation}
where $y \in \mathbb{R}^{H\times W\times2D}$.\\

Additionally, fusing Concatenation fusion leaves the task of defining correspondence to subsequent layers, which in our case is a fully-connected layer \cite{chen2017deeplab}, intended to learn the suitable filters that weigh the layers.

\subsubsection{Fusion Layers}
The authors in \cite{feichtenhofer2016convolutional} report that fusing the streams at different convolution layers has approximately the same number of parameters, and also report that fusing in the later convolution (ReLU5 in the case of \cite{feichtenhofer2016convolutional}) layers yields the best accuracy.
Fusing the streams before the SoftMax layer results in the encoded information from all three streams being passed passed into a fully connected layer, which is responsible for fusing the information from the temporal, spatial, and object-level stream, to generate a classification, as well as a confidence score associated with it, which is basically the confidence of the prediction. Another observation was that pooling of abstract convolution features over Spatio-temporal neighborhoods further boosts the performance of the model, while also maintaining its efficiency. \par

\subsection{Real-time Application - Frame Buffer}

\begin{figure}[b]
    \centering
    \captionsetup{justification=centering}
    \includegraphics[width=7cm]{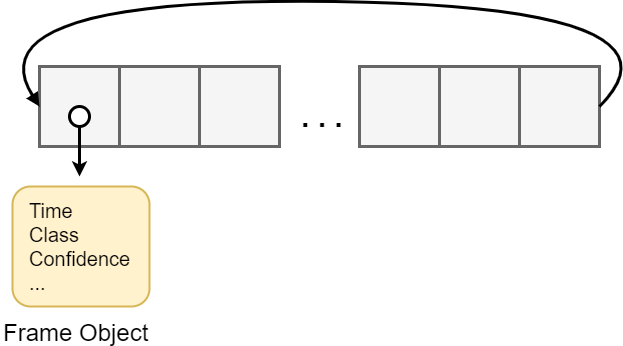}
    \caption{Representation of the Frame Buffer containing Frame objects with information not limited to: i) the timestamp of the frame, ii) predicted classification, iii) confidence score of the prediction (optional).}
    \label{fig:framebuffer}
\end{figure}

To make the real-time application of action recognition a more feasible solution, we have also made our RTAR (Real-time Action Recognition) framework publicly available. The framework uses the three-stream algorithm in a real-time system that revolves around the idea of a circular frame buffer. A frame buffer of a fixed size depending on the number of frames per second of the video stream is created. This frame buffer is circular, which means \textit{'frame object slots'} are recycled using the first come first serve scheduling algorithm.

This scheduling ensures that the most recent frames are always in the buffer, while the oldest frames are removed. The frame buffer is polled at regular intervals (once every 0.5 seconds by default in our framework), to provide a real-time classification of the action currently occurring in the video, based on a majority classification of all frame objects present in the frame buffer at the time, taking into account the \textit{threshold confidence}.
This entire system is designed to be generic and could be extended to get real-time feedback for any sort of video data, or video stream, provided it is trained on the action classes in consideration.

\section{The Hand Wash Dataset}
Typically, most publicly available action recognition datasets are created for generic actions performed day-to-day or are actions of people playing a particular sport, for example, UCF-101 \cite{soomro2012ucf101}, UCF Sports \cite{rodriguez2008action}, KTH \cite{schuldt2004recognizing} and others. \par

The actions performed in these datasets involve large movements from frame to frame, and each of these actions are very unique to each other in the sense that they are performed in widely different settings and environments. For example, the action of applying make-up is vastly different from a field hockey penalty, which are both classes in UCF-101. \par

Further, while these datasets provide a great way to benchmark algorithms, they do not have data that can help in training models needed to solve for a domain-specific problem like identifying armed robbers in a gathering or identifying intricate hand signs in the sign language, that are fine-grained classes, with very little difference between actions of different classes. \par

The requirement is of several videos specific to the problem being solved such that the model has enough \textbf{relevant} data to be able to accurately identify not just any action, but the specific actions the end-user is looking for. \par

When it comes to identifying each step involved in the hand wash procedure as prescribed by the World Health Organisation \cite{boyce2002guideline, pittet2004hand, pittet2009world}, small changes in hand position and changes in the environment need to be taken into account. Since the magnitude of these movements is very small, other action recognition algorithms have a hard time accurately identifying a change from one step to the next. Additionally, current datasets, and not very representative of a potential real-world application of action recognition, such as fixed camera position, with a mostly static background, and adequate training data. Hence, the importance of creating a new dataset was further highlighted. \par

The \textit{Hand Wash Dataset} consists of 292 videos of hand washes created by us (with each hand wash having 12 steps, for a total of 3,504 clips), in different environments to provide as much variance as possible. The variance was important to ensure that the model is robust and can work in more than a few environments. The varied parameters are:
\begin{itemize}
    \item Illumination
    \item Background
    \item Source Camera Position
    \item Field of view
    \item Individuals performing the hand wash (actions)
\end{itemize}
    These parameters were specifically chosen because the \textit{Hand Wash Dataset} intends to simulate the real-world constraints of a potential application of an action recognition solution such as: fixed camera position, real-time feedback, varying illumination, static background and applied to one domain-specific fine-grained action task.
\begin{figure*}
    \centering
    \captionsetup{justification=centering}
    \includegraphics[height = 7cm]{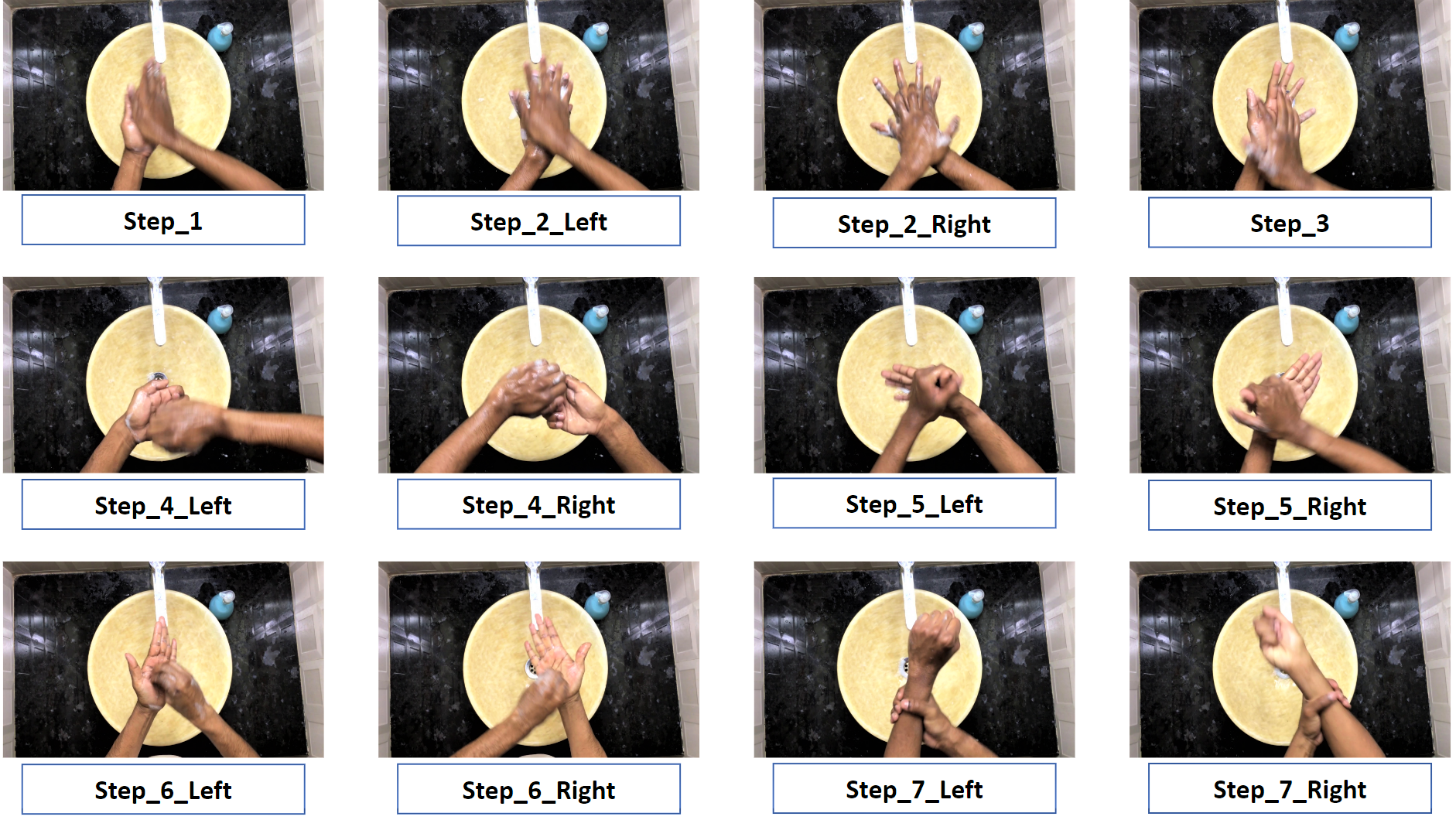}
    \caption{The 12 action classes of the Hand Wash Dataset. Each action class is represented by single sample Frame from a single hand wash clip to maintain uniformity.}
    \label{fig:actionclasses}
\end{figure*}
The next step was to identify each step involved in the procedure. Hence, the original 7 Steps prescribed by the WHO were further broken down into 12 actions which take into account every possible action that may be performed when a person washes their hands. The action classes are :
\newline
    1. \textit{Step\_1}\tab 7. \textit{Step\_5\_Left}
\newline
    2. \textit{Step\_2\_Left}\tab 8. \textit{Step\_5\_Right}
\newline
    3. \textit{Step\_2\_Right}\tab    9. \textit{Step\_6\_Left}
\newline
    4. \textit{Step\_3} \tab          10. \textit{Step\_6\_Right}
\newline
    5. \textit{Step\_4\_Left}\tab     11. \textit{Step\_7\_Left}
\newline
    6. \textit{Step\_4\_Right}  \tab  12. \textit{Step\_7\_Right} 
    \newline
    \newline
The primary difference between the \textit{Hand Wash Dataset} and publicly available datasets such as UCF-101 are:
\begin{enumerate}
    \item The differences between different annotated classes are minimal and fine-grained, (for instance, classes Step 1 and Step 3 as shown in Figure. 1) as compared to UCF-101 or HMDB-101.
    \item The position of the source camera capturing the video is static.
    \item Most of the public datasets available for benchmarking have a large number of classes. The \textit{Hand Wash Dataset} has only 12 very similar annotated classes, as opposed to the 101 vastly distinct classes of UCF-101.
\end{enumerate}
\subsection{Video Clip Groups}
The video clips of each hand wash are first divided into the respective step (or class) out of the 12 possible steps, and then grouped by similar background. Within each of these groups, the videos are varied as much as possible by varying the position of the hands, illumination, field of view, and individuals performing the hand wash.

All the videos used in the \textit{Hand Wash Dataset} were recorded by us, while performing the various steps of the WHO recommended hand wash procedure. All the videos recorded are recorded at a fixed frame rate of 30 FPS, and a fixed resolution of 1920$\times$1080. All audio data has been removed, and the videos are saved in the \textit{.avi} format.
Table 1. summarizes all the details of the \textit{Hand Wash Dataset}.
\begin{table}[h]
  \centering
    \begin{tabular}{| l | l |}
    \hline
    Classes & 12 \\ \hline
    Clips & 3,504 \\ \hline
    Mean Clip Duration & 11.8s \\ \hline
    Frame Rate & 30 FPS \\ \hline
    Resolution & 1920$\times$1080 \\ \hline
    Clips per Group & 10-20 \\ \hline
  \end{tabular}
  \caption{Mean Classification Accuracy for all three models on the public datasets}
  \label{tab:1}
\end{table}
\subsection{Naming Convention}
The compressed version of the dataset is publicly available, and can be found at \cite{handwashdataset2019sample}. This includes 12 folders, each containing the clips of one action class. The name of each video clip follows the following format:
\begin{center}
    $HandWash\_\mathbf{X}\_A\_\mathbf{Y}\_G\_\mathbf{Z}.avi$
\end{center}

where, \\
$\mathbf{X}$ represents the hand wash Number of the video, which is fixed to 3-digit length.\\
$\mathbf{Y}$ represents the action class being performed in the video clip, which is fixed to 2-digit length. \\
$\mathbf{Z}$ represents the background group of the video, which is fixed to 2-digit length.\\

For instance, a video clip labeled \texttt{HandWash\_047\_A\_07\_G\_03.avi} corresponds to the Action Class 7 (class mapping is provided with the dataset), from background group 3, and an overall video ID (or hand wash ID) of 47. 
\begin{table*}[h]
  \centering
  \captionsetup{justification=centering}
  \begin{tabular}{| l | c | c | c |}
    \hline
    \textbf{Model} & \textbf{UCF-101} & \textbf{HMDB-51} & \textbf{\#Parameters} \cr \hline
    IDT with higher-dimensional FV \cite{peng2016bag} & 87.9\% & 61.1\% &  -- \cr \hline
    C3D \cite{tran2015learning}& 85.2\% & -- & 34.8M \cr \hline
    C3D+IDT \cite{tran2015learning}& 90.4\% & -- & 52M \cr  \hline
    TDD+IDT \cite{wang2015action}& 91.5\% & 65.9\% & 124.2M \cr \hline
    Two-stream model (fusion by SVM) \cite{simonyan2014two}& 88.0\%  & 59.4\% & 98M \cr \hline
    Two-stream model (ConvNet - VGG-16) \cite{feichtenhofer2016convolutional}& 92.7\%  & 64.9\% & 181.42M \cr  \hline
    \textbf{Three-Stream Algorithm} & \textbf{91.9\% } & \textbf{62.5\%} & \textbf{23.6M} \cr
    \hline
  \end{tabular}
  \caption{Performance of various action recognition models in terms of i) accuracy on the public datasets; ii) number of parameters in the model which is a direct measure of the compute-intensity of the model}
  \label{tab:2}
\end{table*}
\begin{table*}[h]
  \centering
  \captionsetup{justification=centering}
  \begin{tabular}{| l | c | c |}
    \hline
    \textbf{Model} & \textbf{1\textsuperscript{st} 20 UCF-101} & \textbf{Handwash Dataset} \\ \hline
    RGB Frames Only (Spatial Information)& 91.9\% & 88.7\% \\
    Optical Flow Only (Temporal Information)& 94.3\% & 82.3\% \\
    HOG Only (Object-level Information)& 82.1\%  &67.2\% \\ \hline
    Three-Stream Architecture & 97.7\% & 95.1\%\\
    \hline
  \end{tabular}
  \caption{Mean Classification Accuracy for all three streams individually on the first 20 classes of UCF-101}
  \label{tab:3}
\end{table*}
\subsection{Training and Testing}

We recommend a test train split as mentioned in the documentation of the dataset, with 880 samples in the test split, and 2,624 samples in the train split. A \texttt{.txt} file with the clips to be included in the test and train splits is included with the dataset. Following the train-test split as mentioned would help ensure consistency of experiments reported on the \textit{Hand Wash Dataset}. The test-train splits are designed in a way to ensure there is no overlap between them.\par

Additionally, we also provide scripts to process the dataset at a user-specified frame sampling rate to generate, and write the Optical Flow, as well as Histogram of Oriented Gradients of the frames to disk.

\section{Model Evaluation and Results}
\subsection{Datasets} 
To benchmark our proposed algorithm, two of the largest publicly available datasets were used, namely: UCF-101 and HMDB-51. \par
The UCF-101 \cite{soomro2012ucf101} dataset is a large publicly available dataset that contains videos of various actions sourced from YouTube, with 101 different annotated classes. The HMDB-51 \cite{kuehne2011hmdb} dataset is sourced mainly from various movies, and also from the Prelinger archive, YouTube and Google videos, with 6,849 videos with 51 annotated classes. Both the aforementioned datasets are split into three splits. To obtain an accurate evaluation, the average classification accuracy for all the splits is considered.

Additionally, to focus on fine-grained action recognition in videos, our proposed model was also trained on the \textit{Hand Wash Dataset} introduced in this paper, and performed in real-time on a computer with an Intel i5-6500 processor, and no hardware acceleration.

\subsection{Evaluation}
To evaluate the performance of our three-stream architecture fairly, the results of the three fused streams are compared with the spatial stream, the temporal stream, and the object information stream individually. The evaluation metric used is mean classification accuracy.

All three models were pre-trained on the ImageNet dataset (ILSVRC2012) \cite{deng2009imagenet}, and further trained on the respective datasets (UCF-101 and HMDB-51). 

For evaluation of all the models' performance on the \textit{Hand Wash Dataset}, all three models are trained on the \textit{Hand Wash Dataset}, again pre-trained on the ImageNet dataset. This approach resulted in a better performance as compared to training entirely on the \textit{Hand Wash Dataset} entirely from scratch.

\subsection{Results}

As seen in Table 2, the results of the three-stream algorithm on the UCF-101 dataset are impressive, especially when compared to other commonly used architectures in the domain of action recognition, keeping in mind the fact that the three-stream algorithm, works in real-time on common hardware, with only 23.6 million trainable parameters. \par

An additional experiment we performed was to run our model on 20 fine-grained classes of the UCF-101 dataset. These classes all belong to the same action class (closely related actions), and is far more aligned with real-world constraints. The results of this experiment are seen in Table 3, and it is very evident that the fusion of all three streams gives better results than just using each of the streams individually. This is primarily because of the presence of the subsequent fully-connected layer, responsible for learning abstract convolution features over Spatio-temporal neighborhoods, with object-level information.

\section{Reproducible Research}
In the spirit of reproducible research, we intend to make both the work done, and the entire data set used to train the models for the proof of concept publicly available \cite{kovacevic2007encourage}. All the experiments and work done in this paper is implemented using TensorFlow \cite{abadi2016tensorflow} and we have made a representative sample of the code and data available at present \cite{handwashdataset2019sample, rtarcode2019sample}. Further, the entire code will also be made available and a link to the compendium included in the paper, should the paper be accepted. We would be happy to address any concerns or respond to reviewer questions in this regard.

\section{Conclusion}
A new three-stream algorithm for action recognition in videos is proposed, that incorporates spatial, temporal, and object information using concatenation fusion layers to fuse the three streams, to keep the number of trainable parameters low. This algorithm is also very efficient and can be used in real-time systems, running on commonplace hardware. This is possible as it significantly decreases the number of trainable parameters, yet manages to return impressive results. To confirm the accuracy of the mode it is benchmarked on the UCF-101 and HMDB-51 datasets showing that it can outperform several existing algorithms created to solve a variety of action recognition problems. \par
The \textit{Hand Wash Dataset} was created to provide a dataset that is aligned towards real-world constraints and a potential application of action recognition in real-time such as real-time response to a CCTV stream, automation in a manufacturing line, lip reading, sign-language classification, to name a few. 

{\small
\bibliographystyle{ieee_fullname}
\bibliography{egbib}
}
\end{document}